\documentclass[11pt]{article}

\usepackage[final]{acl}

\usepackage{latexsym}

\usepackage[T1]{fontenc}

\usepackage[utf8]{inputenc}

\usepackage{microtype}

\usepackage{inconsolata}

\usepackage{graphicx}
\usepackage{booktabs}
\usepackage{amsmath}
\usepackage{multirow}
\usepackage{CJKutf8}
\usepackage[T2A,T1]{fontenc}
\usepackage[russian, english]{babel}
\usepackage{times}
\usepackage{caption}

\title{Accommodation Goes Both Ways: Studying Linguistic Convergence Between Humans and Language Models}

\author{Terra Blevins \\
  Khoury College of Computer Sciences\\
  Northeastern University, Boston, MA \\
  \texttt{t.blevins@northeastern.edu}}

\begin{document}
\maketitle
\begin{abstract}
As LLMs become increasingly integrated into daily life, understanding how their presence will shape human linguistic behavior is an open question.
We present a large-scale study of linguistic convergence in human-LLM dialogue, examining how humans and LLMs accommodate each other's linguistic style during multi-turn conversations. Using an asymmetric convergence metric on WildChat, a corpus of real-world ChatGPT transcripts, we find that while LLMs significantly overconverge toward their users on both function word and open-class features across eight languages, human convergence rates in this setting are broadly consistent with human-human baselines. These findings suggest that accommodation in human-LLM dialogue is asymmetric: while LLMs dramatically overfit to their users' style, humans linguistically accommodate LLMs no differently than they would another person.
\end{abstract}

\section{Introduction}
Large language models (LLMs) can produce text indistinguishable from human language in dialogue settings, passing the Turing test \citep{jones2025large}. As these models are increasingly deployed in professional and personal settings, LLMs become new linguistic actors in natural language settings that users can interact with (almost) as naturally as they would another person. Yet despite their growing presence, little is known about how interacting with LLMs shapes the linguistic behavior of their human interlocutors.

We approach this problem through the lens of \textit{linguistic accommodation}, the tendency of speakers to adapt their language based on their interlocutor's over the course of a conversation. This linguistic adaptation is largely unconscious and manifests across a wide range of linguistic features, from phonology and syntax to lexical choices \citep{giles1991accommodation}, and humans have been observed exhibiting some accommodation behaviors to facilitate communication with dialogue systems \cite{brennan1996lexical}. In this work, we build on prior computational methods for studying accommodation in dialogue systems \citep{ireland2011language, danescu2011chameleons} to analyze whether the language of people and models \textit{converges} during human-LLM interactions.

We use this framework to ask:

\noindent \textbf{R1}: Do LLMs converge to the style of their users over the course of a multi-turn conversation?

\noindent \textbf{R2}: Do human users converge to the style of their LLM interlocutor? 

\noindent \textbf{R3}: Does human accommodation of LLMs differ from their accommodation of human interlocutors?

\textbf{R1} is partially addressed in prior work: \citet{kandra2025llms} test whether two LLMs syntactically converge over time, and \citet{blevins2026language} evaluate whether LLMs stylistically accommodate their context when asked to replace turns in a preexisting conversation. However, these settings are both synthetic evaluations using model-model interactions and turn-replacement prompts, respectively; no studies have examined the per-turn coordination of both humans and LLMs in a realistic chatbot setting across multiple languages. Moreover, the human side of these interactions remains underexplored, with open questions about if and how users adapt their language when conversing with LLMs.

We examine these questions with WildChat \citep{zhao2024wildchat}, a corpus of human-LLM conversations drawn from the transcripts of opt-in ChatGPT interactions, to compare the accommodation behavior of LLMs and humans across eight languages on multiple forms of lexical convergence, contrasting the human behavior in this setting with that observed in human-human dialogue. Our analysis reveals that LLMs accommodate users significantly more than humans accommodate LLMs, consistent with prior synthetic evaluations \cite{blevins2026language}. Surprisingly, we also find that while users accommodate less than LLMs in WildChat dialogues, human accommodation rates in human-LLM interaction are broadly consistent with human-human dialogue.
This asymmetry holds for all features and languages examined, suggesting that the accommodation of both humans and LLMs in this setting is linguistically and cross-lingually robust.
These findings suggest that humans, at least linguistically, accommodate modern LLMs no differently than they would another person, despite the asymmetric and overaccommodating behavior exhibited by the LLM.

\section{Related Work}
Accommodation is a well-documented phenomenon where speakers adapt their language to their interlocutor, manifesting across features from phonology to lexical choice \citep{giles1991accommodation, brennan1996conceptual}. We focus on linguistic convergence, where a speaker's style becomes more similar to their interlocutor's over time \cite{niederhoffer2002linguistic}. \citet{ireland2011language} and \citet{danescu2011chameleons} develop computational methods for measuring convergence in large corpora, enabling the study of accommodation at scale, and \citet{ward2007automatically} extends this to lexical entrainment, showing that accommodation on content words can be measured automatically. These methods have been broadly applied to human-human dialogue settings \cite[][i.a.]{mukherjee2012analysis, bawa2018accommodation, berdivcevskis2023you}.

Recently, these methods have been used to study LLM accommodation, albeit in artificial settings. \citet{kandra2025llms} find that LLMs syntactically converge in model-to-model interactions, and \citet{blevins2026language} show that LLMs accommodate their context when completing turns in existing human conversations. We extend these findings by studying accommodation in preexisting human-LLM interactions, enabling a direct comparison of LLM and human accommodation in this setting. Concurrently to our work, \citet{chen2026accommodates} apply a bidirectional within-vs-between design to English GPT-4o conversations and find a similar LLM-User asymmetry. We corroborate these prior findings and show that LLMs also overconverge in this deployed, realistic setting. 

A parallel line of work on computers as social actors shows that people unconsciously apply social rules to computers \citep{nass1994computers, nass2000machines}. However, the case for user accommodation of computers is mixed: \citet{brennan1996lexical} finds that users entrain to computer-generated referring expressions, but recent work on human-model interaction indicates that users adopt different linguistic behaviors with LLMs \cite{bhatt2021detecting, zhang2025mind}. We directly address this question, finding that human accommodation is broadly stable across LLM- and human-human settings.

\section{Methods and Experimental Setup}
To study how humans and LLMs accommodate each other, we analyze conversations from real-world corpora across multiple languages.

\begin{table}[t]
\centering
\small
\setlength{\tabcolsep}{3pt}
\begin{tabular}{llrrr}
\toprule
\textbf{Dataset} & \textbf{Lang} & \textbf{\# Convos} & \textbf{Avg.\ Turns} & \textbf{\# Tokens} \\
\midrule
WildChat & EN & 5,000 & 6.0 & 12.8M \\
 & ES & 5,000 & 4.8 & 8.8M \\
 & FR & 5,000 & 5.0 & 9.4M \\
 & IT & 2,482 & 3.7 & 4.5M \\
 & PT & 4,811 & 3.5 & 5.7M \\
 & RU & 5,000 & 6.0 & 7.3M \\
 & TR & 2,563 & 3.7 & 2.9M \\
 & ZH & 5,000 & 6.0 & 8.4M \\
\midrule
DailyDialog & EN & 5,000 & 4.6 & 652K \\
Ubuntu & EN & 5,000 & 6.0 & 1.2M \\
\bottomrule
\end{tabular}
\caption{Dataset statistics for all experimental settings. WildChat represents human-LLM dialogue, while DailyDialog and Ubuntu cover human-human conversations.}
\vspace{-4pt}
\label{tab:statistics}
\end{table}

\begin{table*}[t]
\centering
\small
\setlength{\tabcolsep}{4pt}
\begin{tabular}{lrrrrrrrrrrr}
\toprule
\textbf{Setting} & \textbf{NOUN} & \textbf{LIWC:} & \textbf{ppron} & \textbf{ipron} & \textbf{article} & \textbf{conj} & \textbf{prep} & \textbf{auxvb} & \textbf{adverb} & \textbf{negate} & \textbf{quant} \\
\midrule
WildChat LLM      & .442 & .068 & .086 & .051 & .041 & .035 & .022 & .033 & .063 & .213 & .067 \\
WildChat User     & .141 & .035 & .048 & .031 & .028 & .021 & .018 & .022 & .032 & .081 & .033 \\
\quad WildChat $\Delta$    & .301 & .033 & .038 & .020 & .013 & .014 & .004 & .011 & .031 & .132 & .034 \\
\midrule
DailyDialog       & .132 & .036 & .028 & .048 & .046 & .038 & .030 & .015 & .039 & \underline{$-.005$} & .084 \\
\midrule
Ubuntu User       & .104 & .022 & .020 & .022 & .024 & .025 & .019 & .013 & .018 & \underline{$.012$} & .040 \\
Ubuntu Asst.      & .103 & .020 & .019 & .021 & .025 & .024 & .019 & .013 & .016 & \underline{$.011$} & .038 \\
\bottomrule
\end{tabular}
\caption{Convergence scores for WildChat (EN), DailyDialog, and Ubuntu on averaged NOUN and LIWC scores, as well as each function word class in LIWC. All values are significant at $p < 0.05$ unless underlined. $\Delta$ = WildChat LLM $-$ User. DailyDialog scores are averaged across both speakers.}
\vspace{-4pt}
\label{tab:results_en}
\end{table*}

\subsection{Data} 
\label{sec:data}
Our primary analysis is conducted on Wildchat \cite{zhao2024wildchat}, a corpus of approximately one million real-world conversations between human users and ChatGPT (GPT-3.5-Turbo and GPT-4) spanning many topics and languages, making it well-suited for studying organic human-LLM interactions. For our study, we filter these conversations to those with 2 to 10 (human, LLM) turn pairs. We then sample up to 5,000 conversations per language across 8 languages (English, French, Spanish, Portuguese, Italian, Russian, Chinese, and Turkish), using stratified sampling by turn count to ensure a range of conversation lengths in our sample. Included languages must occur frequently ($\sim2.5k$ or more conversations after filtering) and have the language-specific resources needed to calculate convergence features.

As a counterpoint to our study of LLM-human conversations, we also consider speaker behavior in human-human dialogue settings. Given the inherent paired structure of chatbot conversations, we focus on dyadic settings, or conversations with two speakers, and consider two English datasets that meet this criteria: DailyDialog \cite{li2017dailydialog}, a widely used dialogue dataset of everyday conversations, and the original conversations in the Ubuntu Dialogue Corpus \cite{lowe2015ubuntu}, a corpus of technical support conversations from the Ubuntu IRC channel that provides a task-oriented human-human baseline representing a common LLM use case.\footnote{Due to their release dates, there is little risk of machine-generated content in either corpus.} We follow the same preprocessing steps as on WildChat for human-human baselines, requiring all turns between speakers to be paired. Table \ref{tab:statistics} provides our data statistics after filtering.

\subsection{Convergence Metrics}
\label{sec:methods}
Unlike prior work on accommodation in LLMs \cite[e.g.,][]{blevins2026language}, we study the separate behaviors of humans and LLMs in interactive chatbot settings, thus requiring an asymmetric measure to disentangle the effect of conversational partners on each other. We adopt the convergence metric from \citet{danescu2011chameleons}, which measures the degree to which one speaker accommodates another on a binary feature \texttt{t}. For exchanges where $A$ initiates and $B$ responds, let $a^t$ and $b^t_{\rightarrow a}$ be indicator variables for $A$'s utterances and $B$'s replies exhibiting feature $t$. Then the coordination of $B$ to $A$ is defined as $\text{Conv}(B \rightarrow A, t)$:
\begin{equation} P(b^t_{\rightarrow a} = 1 \mid a^t = 1) - P(b^t_{\rightarrow a} = 1)\end{equation}
Positive scores indicate $B$ uses $t$ above their baseline rate when $A$ does (\textit{convergence}); negative scores indicate \textit{divergence}. Conv is averaged across all turns in the sample, unless otherwise stated.

We measure convergence on two feature types:

\textbf{LIWC function words} are frequent, largely unconscious markers that are sensitive to accommodation effects in human dialogue \citep{niederhoffer2002linguistic}. Following \citet{danescu2011chameleons} and \citet{ireland2011language}, we measure convergence on function word categories from the Linguistic Inquiry and Word Count (LIWC) lexicon \citep{chung2012linguistic}. We use language-specific LIWC 2007 dictionaries (Appendix \ref{app:experiment}) for the languages in our study, reporting per-category and mean scores.\footnote{Function word coverage varies slightly by language due to differences in grammar and dictionary versions — for example, articles are absent in Chinese and Turkish.}

\textbf{NOUN lemmas} reflect lexical entrainment \cite{brennan1996conceptual}, or the tendency of speakers to adopt each other's specific vocabulary choices when discussing a concept. Following \citet{ward2007automatically}, we extract the 100 most frequent noun lemmas per corpus after filtering to nouns with at least one WordNet synonym \citep{fellbaum1998wordnet} when possible to ensure each lemma represents a lexical choice.\footnote{WordNet filtering is available for EN, FR, ES, PT, IT, and ZH via NLTK OMW 1.4 \cite{bond2013linking}; RU and TR use frequency-only filtering due to missing coverage.} We report mean convergence score across selected nouns in the vocabulary; Appendix Table \ref{tab:vocab} reports which lemmas are selected in each setting.

\section{Results}
Table \ref{tab:results_en} reports our experimental results on English WildChat, as well as the two human-human baselines. Generally, humans and LLMs both exhibit significant convergence over random chance. Furthermore, we observe that:

\paragraph{Human speakers accommodate LLMs and humans similarly} We see broadly consistent convergence scores across the human speaker settings of \textit{WildChat User}, \textit{DailyDialog} speakers, and both \textit{Ubuntu} settings (which are split due to the different roles of the user asking for help and the expert assisting the user): mean LIWC convergence scores for human speakers range from .022--.036 across all three settings, and noun convergence scores from .104--.141. This indicates that not only are speakers generally consistent across different dialogue settings and roles in our chosen corpora, but also that their behavior towards LLM interlocutors is similar on the considered metrics.  One exception where speakers accommodate LLMs \textit{more} is on negation: \textit{WildChat} users show notably higher negation convergence (.081) than the non-significant scores of \textit{DailyDialog} (-.005) or \textit{Ubuntu} (.012), which is potentially driven by the (much) larger convergence of LLMs on the same word class (.213). %

\paragraph{LLMs overconverge to their user} 
We find that LLMs significantly \textit{overconverge} towards their users, both relative to the users they are speaking with and to human-human accommodation rates in our two baseline corpora. This is consistent with prior work \cite{blevins2026language} and further shows that LLM mirroring of user style is excessive relative to human behavior even in realistic, deployed settings. Specifically, mean convergence of LLMs on LIWC is roughly twice the user rate ($.068$ vs $.035$), with relative increases of LLM scores over users on individual features ranging from $22\%$ on prepositions to $164\%$ on negating function words (an outlier that contrasts sharply with the minimal negation convergence seen in both human-human baselines). LLM convergence on nouns is even more pronounced: ChatGPT achieves a mean noun convergence of $.442$, compared to scores of $.141$ for users in WildChat and $.104$ to $.132$ across the human-human baselines. As LLM noun convergence occurs at over three times the user rate, this indicates ChatGPT is even more sensitive to user lexical choice beyond function words.

\begin{figure}[t]
    \centering
    \includegraphics[width=0.9\linewidth]{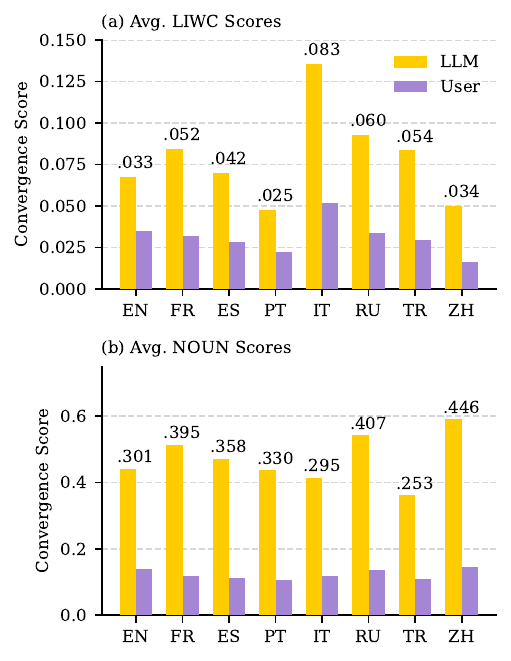}
    \caption{Mean (a) LIWC and (b) NOUN scores for LLMs and users across eight languages in WildChat. LLMs consistently converge more than users across all languages; the reported values indicate $\Delta$ = LLM - User.}
    \vspace{-4pt}
    \label{fig:crosslingual}
\end{figure}

\paragraph{LLM convergence is robust across languages}
The observed asymmetry between user and LLM convergence in WildChat holds across the eight considered languages (Figure \ref{fig:crosslingual}).\footnote{Full per-language results are in Appendix Table \ref{tab:results_multilingual}.} Despite typological differences across the eight languages and (presumed) variation in ChatGPT training data coverage, the asymmetry is remarkably consistent. ChatGPT's average LIWC convergence scores range from $.048$ (PT) to $.135$ (IT), while user scores remain stable between $.016$ (ZH) and $.052$ (IT); relatively, ChatGPT converges between $95\%$-$213\%$ above user rates across languages. Notably, English, which is often the best-resourced language for LLM training, falls in the middle of this range rather than at the top or bottom for any individual feature, suggesting the asymmetry is not simply a function of model quality on a given language. These same patterns hold for NOUN convergence, with LLM scores exceeding user rates by $214\%-330\%$ across all languages; this increase in convergence on nouns also reinforces the finding that LLMs are particularly responsive to user choices of open-class lexical items.

\begin{figure*}
    \centering
    \includegraphics[width=1\linewidth]{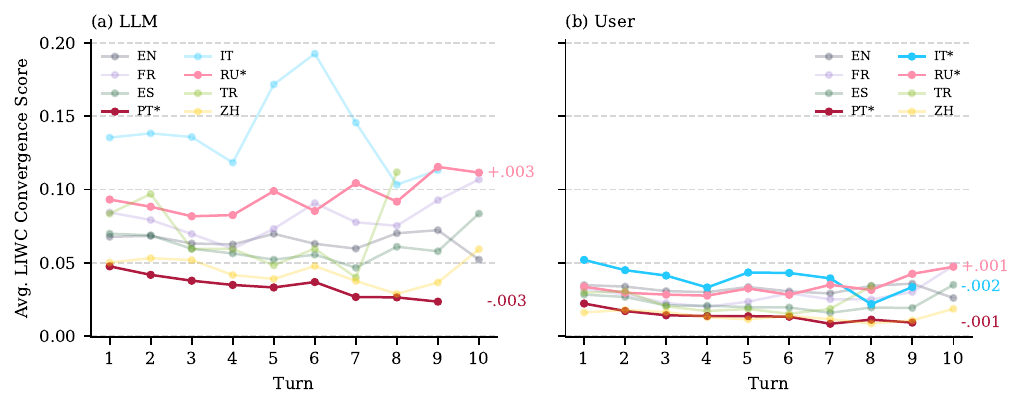}
    \captionsetup{skip=6pt}
    \caption{Mean LIWC convergence scores per turn position for LLMs and users across eight WildChat languages. Significant trends are indicated with an $*$, saturated lines, and reported correlation value($p < 0.05$).}
    \vspace{-4pt}
    \label{fig:liwc_perturn}
\end{figure*}

\paragraph{Per-turn convergence is consistent over time}
While our prior experiments consider the overall convergence rates in human-LLM dialogue, it remains an open question whether these rates change over the course of conversations.
We therefore perform a temporal analysis to compare the per-turn convergence rates of users and LLMs; Figure \ref{fig:liwc_perturn} reports the mean LIWC scores at each turn in the conversations, while Appendix Figure \ref{fig:noun_perturn} shows the corresponding NOUN results.\footnote{We do not include cases where there are $<100$ utterances (e.g., on TR turns 9 and 10, as well as IT, PT turn 10).} We also fit a linear regression of convergence score on turn position for each setting to test whether scores change significantly over time, finding that 10 of 32 cases show a significant linear trend ($p<0.05$), indicated with their slope values in each figure.

Overall, we find no consistent patterns, with most fluctuations appearing as noise or barely significant (only five cases remain so under a stricter p-value of $0.01$). Significant slopes are also distributed across both positive and negative trends: RU users and models increasingly converge on LIWC while PT speakers both diverge, suggesting that even real trends in the data reflect language-specific variation rather than a systematic tendency for either speaker type to accommodate more or less over time. Human-human baselines are similarly flat on LIWC across turns, with no significant trends in either DailyDialog or Ubuntu. However, DailyDialog NOUN scores show a significant decreasing trend ($p<.01$), which may be due to the domain (language learners) or increased data sparsity on later turns.

These findings somewhat contrast with \citet{chen2026accommodates}, who study English GPT-4o conversations ($n=1319$) and report progressive user convergence on personal pronouns across turns. The discrepancy may be due to methodological differences: they measure whether speakers become increasingly similar to their own conversational partner relative to random conversations, whereas we measure local turn-by-turn coordination.\footnote{Note that stable coordination under our metric does not preclude the increasing similarity that \citet{chen2026accommodates} report. If speakers consistently coordinate throughout a conversation, their language will cumulatively converge even without an increase in the per-turn coordination rate.} It may also reflect a setting-specific case, given that this effect is limited to personal pronouns and does not replicate across their other features, or the broader set of languages and corpus sizes examined here.

\section{Conclusion}
We study linguistic convergence in human-LLM dialogue, finding that LLMs significantly overconverge toward their users on both function and open-class lexical features, while human accommodation rates remain stable across LLM and human-human settings. This asymmetry holds consistently across eight languages and both feature types, suggesting it is a robust property of human-LLM interaction.

Notably, the conclusions we draw about human behavior differ from \citet{bhatt2021detecting}: they find that humans accommodate bots differently than other humans, which they attribute to their systems often producing incoherent responses that disrupt natural conversation, unlike the more coherent outputs of modern LLMs. Based on the consensus that linguistic accommodation is a largely unconscious process \citep{giles1991accommodation, pickering2004toward}, we hypothesize that LLM fluency acts as a bottleneck: once the model produces sufficiently coherent language in dialogue, these processes occur as they do in human-human dialogue, regardless of the interlocutor's nature. As model quality improves and LLMs are increasingly integrated into everyday communication, understanding how humans subconsciously treat these systems will be critical for anticipating the long-term effects of LLMs on natural language.

\section*{Limitations}
Our analysis is limited to a single LLM provider, as WildChat conversations are drawn exclusively from OpenAI's ChatGPT (GPT-3.5-Turbo and GPT-4); thus, it is unclear whether the observed overconvergence generalizes to other LLM families, though it is consistent with prior work \citep{blevins2026language, kandra2025llms}. Relying exclusively on ChatGPT introduces the confound of post-training alignment, which we do not attempt to disentangle from the base pretraining objective but leave for future work. Additionally, although we study eight languages, our human-human baselines are English-only, which limits the cross-lingual scope of our findings. LIWC dictionary coverage also varies across languages, with some categories absent entirely for certain languages (e.g., RU, TR, and ZH all do not have an article category), which may affect cross-lingual comparability of our per-feature LIWC results. WordNet's OMW coverage is similarly limited by its lack of support for RU and TR, which may affect our NOUN results on those languages.
Beyond WordNet, our noun lemma vocabularies are still corpus-specific and may reflect domain biases of their underlying data (for example, the prevalence of technical terms in the vocabulary for Ubuntu), which could skew noun convergence scores based on the conversational setting. Future work should examine convergence behavior across a broader range of LLM families and deployment settings and develop human-human dialogue baselines in languages beyond English to fully enable a cross-lingual comparison of human and LLM accommodation behavior.

\bibliography{custom}

\appendix

\section{More Experimental Details}
\label{app:experiment}
We use LIWC dictionaries to obtain inventories of words that fall into each function word class measured as a proxy for linguistic convergence in prior work \cite{ireland2011language, danescu2011chameleons}: personal pronouns (ppron), impersonal pronouns (ipron), articles, conjunctions (conj), prepositions (prep), auxiliary verbs (auxvb), adverbs, negations (negate), and quantifiers (quant). In non-English experimental settings, we use language-specific dictionaries that were constructed in the vein of LIWC-2007: Spanish \cite{ramirez2007psicologia}; French \cite{piolat2011french}; (Brazilian) Portuguese \cite{balage2013evaluation}; Italian \cite{agosti2007italian}; Russian \cite{kailer2011russian}; Turkish \cite{boyd2022development}; and Simplified Chinese \cite{lyu2022psychological}. We note that we do not include other languages, such as German, whose dictionaries are based on different versions of LIWC, as these have lower overlap with the word-class categories considered in this paper.

We report the lemma inventories used to calculate NOUN scores for all settings in Table \ref{tab:vocab}. While these inventories are automatically extracted from the data, they are mostly reasonable for their target language; however, the Turkish vocabulary contains some English tokens (e.g., \textit{the}, \textit{and}, \textit{for}), likely due to code-switching in WildChat Turkish conversations and the lack of WordNet filtering for Turkish, which in other languages serves to remove low-quality noun candidates. To obtain our tokens and noun lemmas for our analysis, we process all text using a language-appropriate SpaCy pipeline \cite{honnibal2020spacy}; we use a community-hosted pipeline for Turkish.\footnote{\url{https://huggingface.co/turkish-nlp-suite/tr_core_news_md}.}

We use the following data and software resources: WildChat \cite{zhao2024wildchat} (ODC-BY license), DailyDialog \cite{li2017dailydialog} (CC BY-NC-SA 4.0), spaCy \cite{honnibal2020spacy} (MIT license), WordNet \cite{fellbaum1998wordnet} (Princeton WordNet License), and OMW 1.4 (CC BY 4.0). LIWC 2007 dictionaries \cite{chung2012linguistic} are used under a purchased academic license and are not redistributed. The Ubuntu Dialogue Corpus \cite{lowe2015ubuntu} is derived from publicly available Ubuntu IRC chat logs; while commonly used, no explicit license is associated with the dataset. Specifically, we use WildChat, which has been de-identified using Microsoft Presidio and custom rules to identify and remove PII across various data types in English, Chinese, Russian, French, Spanish, German, Portuguese, Italian, Japanese, and Korean. DailyDialog and the Ubuntu Dialogue Corpus are widely used academic datasets that do not contain identifying information. We do not collect any new data in this work.

\begin{table}[b!]
\centering
\small
\setlength{\tabcolsep}{4pt}
\begin{tabular}{llrrrr}
\toprule
 & & \multicolumn{2}{c}{\textbf{LIWC}} & \multicolumn{2}{c}{\textbf{NOUN}} \\
\cmidrule(lr){3-4} \cmidrule(lr){5-6}
\textbf{Setting} & \textbf{Speaker} & \textbf{Slope} & \textbf{$p$} & \textbf{Slope} & \textbf{$p$} \\
\midrule
EN & LLM & -0.0006 & 0.425 & 0.0084 & 0.004$^*$ \\
 & User & -0.0003 & 0.375 & -0.0009 & 0.085 \\
FR & LLM & 0.0024 & 0.104 & -0.0052 & 0.495 \\
 & User & 0.0011 & 0.205 & 0.0005 & 0.620 \\
ES & LLM & 0.0002 & 0.892 & -0.0045 & 0.311 \\
 & User & -0.0001 & 0.901 & -0.0019 & 0.017$^*$ \\
PT & LLM & -0.0027 & 0.000$^*$ & -0.0037 & 0.421 \\
 & User & -0.0014 & 0.001$^*$ & -0.0010 & 0.197 \\
IT & LLM & -0.0017 & 0.679 & -0.0036 & 0.685 \\
 & User & -0.0023 & 0.028$^*$ & -0.0024 & 0.096 \\
RU & LLM & 0.0028 & 0.023$^*$ & 0.0021 & 0.376 \\
 & User & 0.0015 & 0.026$^*$ & 0.0020 & 0.028$^*$ \\
TR & LLM & -0.0012 & 0.788 & -0.0086 & 0.214 \\
 & User & -0.0005 & 0.682 & -0.0021 & 0.165 \\
ZH & LLM & -0.0009 & 0.398 & -0.0529 & 0.000$^*$ \\
 & User & -0.0004 & 0.285 & -0.0103 & 0.003$^*$ \\
\midrule
DD & User & -0.0002 & 0.751 & -0.0118 & 0.003$^*$ \\
Ubuntu & User & -0.0006 & 0.325 & -0.0013 & 0.110 \\
\bottomrule
\end{tabular}
\caption{Linear trend slopes and $p$-values for per-turn convergence scores across all settings. $^*p < 0.05$. DD and Ubuntu report only user/speaker scores since there is no LLM speaker.}
\label{tab:perturn_trends}
\end{table}

\section{Additional Results}
\label{app:results}
\begin{table*}[t]
\centering
\small
\setlength{\tabcolsep}{3pt}
\begin{tabular}{llrrrrrrrrrrr}
\toprule
\textbf{Lang} & \textbf{Speaker} & \textbf{NOUN} & \textbf{LIWC:} & \textbf{ppron} & \textbf{ipron} & \textbf{article} & \textbf{conj} & \textbf{prep} & \textbf{auxvb} & \textbf{adverb} & \textbf{negate} & \textbf{quant} \\
\midrule
\multirow{3}{*}{EN}
  & LLM  & 0.442 & 0.068 & 0.086 & 0.051 & 0.041 & 0.035 & 0.022 & 0.033 & 0.063 & 0.213 & 0.067 \\
  & User & 0.141 & 0.035 & 0.048 & 0.030 & 0.028 & 0.021 & 0.018 & 0.021 & 0.032 & 0.081 & 0.033 \\
  & \quad$\Delta$ & 0.301 & 0.033 & 0.038 & 0.020 & 0.013 & 0.014 & 0.004 & 0.011 & 0.031 & 0.132 & 0.034 \\
\midrule
\multirow{3}{*}{FR}
  & LLM  & 0.514 & 0.084 & 0.019 & 0.225 & 0.039 & 0.032 & 0.031 & 0.066 & 0.045 & 0.252 & 0.051 \\
  & User & 0.119 & 0.032 & 0.012 & 0.045 & 0.034 & 0.015 & 0.025 & 0.029 & 0.018 & 0.082 & 0.029 \\
  & \quad$\Delta$ & 0.395 & 0.052 & 0.007 & 0.180 & 0.005 & 0.016 & 0.006 & 0.037 & 0.027 & 0.170 & 0.021 \\
\midrule
\multirow{3}{*}{ES}
  & LLM  & 0.471 & 0.070 & 0.024 & 0.038 & 0.043 & 0.034 & 0.008 & 0.085 & 0.041 & 0.256 & 0.099 \\
  & User & 0.113 & 0.028 & 0.017 & 0.022 & 0.034 & 0.020 & 0.007 & 0.023 & 0.017 & 0.081 & 0.033 \\
  & \quad$\Delta$ & 0.358 & 0.042 & 0.007 & 0.017 & 0.009 & 0.014 & 0.002 & 0.062 & 0.023 & 0.175 & 0.066 \\
\midrule
\multirow{3}{*}{PT}
  & LLM  & 0.437 & 0.048 & 0.021 & 0.034 & 0.045 & 0.025 & 0.005 & 0.024 & 0.035 & 0.199 & 0.040 \\
  & User & 0.107 & 0.022 & 0.015 & 0.024 & 0.033 & 0.016 & 0.004 & 0.014 & 0.014 & 0.057 & 0.023 \\
  & \quad$\Delta$ & 0.330 & 0.025 & 0.006 & 0.010 & 0.012 & 0.009 & 0.001 & 0.011 & 0.021 & 0.142 & 0.017 \\
\midrule
\multirow{3}{*}{IT}
  & LLM  & 0.414 & 0.135 & 0.060 & -- & 0.043 & -- & 0.025 & -- & -- & 0.212 & 0.337 \\
  & User & 0.119 & 0.052 & 0.028 & -- & 0.030 & -- & 0.018 & -- & -- & 0.069 & 0.115 \\
  & \quad$\Delta$ & 0.295 & 0.084 & 0.033 & -- & 0.013 & -- & 0.007 & -- & -- & 0.143 & 0.222 \\
\midrule
\multirow{3}{*}{RU}
  & LLM  & 0.542 & 0.093 & 0.043 & 0.118 & -- & 0.036 & 0.014 & -- & 0.085 & 0.230 & 0.126 \\
  & User & 0.135 & 0.034 & 0.020 & 0.024 & -- & 0.022 & 0.011 & -- & 0.034 & 0.089 & 0.036 \\
  & \quad$\Delta$ & 0.407 & 0.059 & 0.023 & 0.095 & -- & 0.014 & 0.003 & -- & 0.051 & 0.141 & 0.090 \\
\midrule
\multirow{3}{*}{TR}
  & LLM  & 0.360 & 0.084 & 0.106 & 0.054 & -- & 0.047 & 0.070 & 0.133 & \underline{-0.007} & 0.214 & 0.050 \\
  & User & 0.107 & 0.030 & 0.059 & 0.024 & -- & 0.022 & 0.019 & 0.032 & \underline{-0.005} & 0.064 & 0.021 \\
  & \quad$\Delta$ & 0.253 & 0.054 & 0.048 & 0.030 & -- & 0.025 & 0.052 & 0.101 & -0.002 & 0.150 & 0.028 \\
\midrule
\multirow{3}{*}{ZH}
  & LLM  & 0.590 & 0.050 & 0.202 & 0.031 & -- & \underline{-0.006} & \underline{0.003} & 0.024 & 0.015 & 0.110 & 0.022 \\
  & User & 0.144 & 0.016 & 0.074 & 0.011 & -- & \underline{-0.002} & \underline{0.001} & 0.006 & 0.006 & 0.026 & 0.006 \\
  & \quad$\Delta$ & 0.446 & 0.034 & 0.128 & 0.020 & -- & \underline{-0.004} & \underline{0.001} & 0.018 & 0.009 & 0.083 & 0.017 \\
\bottomrule
\end{tabular}
\caption{Per-language convergence scores WildChat on averaged NOUN and LIWC scores, as well as each function word class in LIWC. All values are significant at $p < 0.05$ unless underlined. $\Delta$ = LLM $-$ User. Cells marked -- indicate categories absent from that language\'s LIWC dictionary.}
\vspace{-10pt}
\label{tab:results_multilingual}
\end{table*}

\begin{figure*}
    \centering
    \includegraphics[width=1\linewidth]{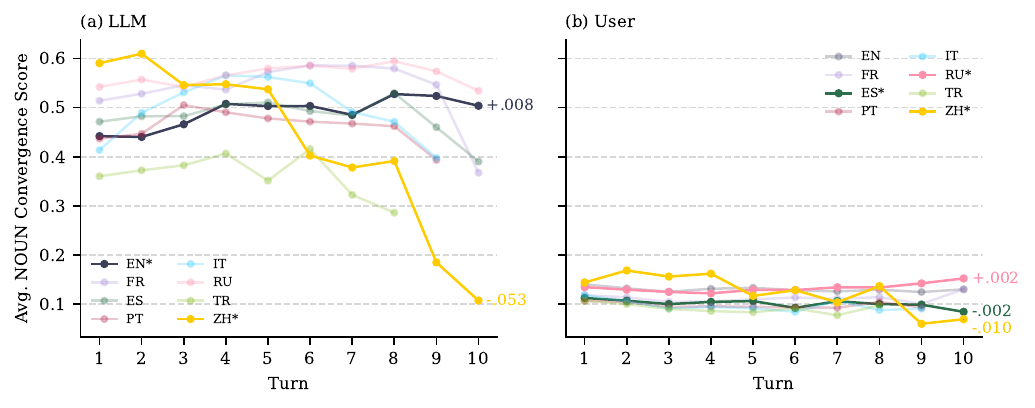}
    \captionsetup{skip=6pt}
    \caption{Mean NOUN convergence scores per turn position for LLMs and users across eight WildChat languages. Significant trends are indicated with an $*$, saturated lines, and reported correlation value($p < 0.05$). One notable outlier is ZH in this setting: both human and LLM speakers converge less over time, with LLMs in particular sharply mirroring their user less on later turns.}
    \label{fig:noun_perturn}
\end{figure*}

Table \ref{tab:results_multilingual} presents the full results for WildChat on all languages and features, while Figure \ref{fig:noun_perturn} reports the per-turn convergence results for WildChat on NOUNs. We also report the full statistical analysis of the per-turn trends (Table \ref{tab:perturn_trends}).

\begin{table*}[t]
\centering
\tiny
\begin{tabular}{lp{0.85\textwidth}}
\toprule
\textbf{Lang} & \textbf{Nouns} \\
\midrule
EN & time, datum, value, user, code, function, system, information, file, name, example, model, number, process, world, way, character, point, game, class, life, power, method, text, language, experience, content, error, task, step, question, product, type, action, change, result, word, ability, day, year, development, level, import, response, event, line, image, people, service, issue, company, research, performance, role, project, student, return, application, part, strategy, case, team, approach, end, term, technology, story, market, skill, element, feature, object, individual, command, input, list, player, message, form, state, challenge, business, group, area, customer, decision, work, source, energy, structure, use, need, impact, resource, script, color, relationship, factor, tool, community \\
\midrule
FR & donnée, système, gestion, fonction, information, entreprise, exemple, besoin, valeur, question, recherche, temps, code, utilisation, jour, projet, application, manière, analyse, processus, utilisateur, pourcent, service, travail, problème, développement, modèle, base, méthode, nom, groupe, sécurité, étude, produit, vie, objectif, ligne, nombre, cas, compte, résultat, place, moment, test, formation, type, communication, expérience, vidéo, compétence, étape, ressource, qualité, personne, erreur, mise, jeu, point, élément, mesure, domaine, réponse, réseau, équipe, image, fichier, niveau, activité, variable, solution, fois, outil, capacité, relation, condition, approche, santé, terme, texte, environnement, risque, rapport, membre, performance, client, rôle, contexte, ensemble, eau, mot, demande, accès, traitement, monde, état, structure, action, décision, situation, forme \\
\midrule
ES & ejemplo, información, persona, manera, sistema, valor, tiempo, proceso, caso, función, vida, forma, parte, palabra, empresa, servicio, problema, tipo, trabajo, respuesta, día, modelo, resultado, nombre, nivel, producto, lugar, desarrollo, relación, calidad, aplicación, texto, red, cuenta, uso, número, capacidad, necesidad, derecho, base, código, seguridad, situación, cambio, proyecto, punto, estudiante, habilidad, imagen, análisis, recurso, contexto, objetivo, equipo, actividad, experiencia, acción, mundo, cliente, opción, pregunta, historia, tema, área, línea, medida, herramienta, conocimiento, momento, error, paso, idea, método, contenido, juego, prueba, comunicación, educación, país, estado, elemento, estudio, color, cantidad, estructura, estrategia, objeto, campo, clave, tabla, resumen, enfoque, investigación, decisión, clase, condición, riesgo, importancia, gestión, lenguaje \\
\midrule
PT & forma, pessoa, informação, exemplo, vida, tempo, dado, mundo, jogo, vez, sistema, valor, história, habilidade, equipe, resposta, ano, função, dia, processo, empresa, palavra, problema, parte, modelo, texto, arquivo, nome, personagem, poder, ponto, experiência, conhecimento, relação, desenvolvimento, área, tipo, recurso, maneira, base, capacidade, necessidade, objetivo, série, uso, resultado, ação, código, trabalho, caso, serviço, gente, número, pergunta, imagem, desafio, opção, usuário, contexto, livro, questão, cliente, meio, coisa, projeto, saúde, ambiente, momento, qualidade, ideia, produto, conteúdo, elemento, atividade, acordo, evento, prática, energia, inteligência, cidade, carro, situação, mudança, acesso, tecnologia, lugar, linha, nível, importância, universo, grupo, segurança, linguagem, conceito, versão, corrida, estratégia, mercado, caminho, tema \\
\midrule
IT & dato, esempio, modo, numero, tempo, codice, sistema, mondo, valore, funzione, vita, parte, modello, caso, base, lavoro, volta, storia, persona, giorno, problema, anno, articolo, attività, gestione, ricerca, metodo, tipo, punto, servizio, ordine, processo, relazione, momento, interno, ambiente, testo, nome, natura, errore, elemento, utente, contesto, uso, oggetto, operazione, casa, prodotto, risposta, livello, spazio, risorsa, stato, struttura, termine, esperienza, condizione, gioco, applicazione, città, forma, fine, forza, sicurezza, rete, controllo, classe, strumento, acqua, campo, import, soluzione, conoscenza, parola, strategia, chiave, comunità, risultato, capacità, posizione, analisi, serie, prezzo, indirizzo, gruppo, sito, cuore, società, lavoratore, domanda, contenuto, scelta, azione, obiettivo, libro, sfida, tabella, situazione, variabile, realtà \\
\midrule
RU & \foreignlanguage{russian}{система, функция, значение, работа, использование, данных, метод, код, время, процесс, информация, пример, файл, человек, элемент, число, год, создание, помощь, развитие, результат, задача, решение, текст, образ, проект, изменение, проблема, строка, возможность, уровень, количество, пользователь, программа, ответ, ошибка, анализ, модель, объект, случай, качество, команда, вопрос, цель, управление, условие, организация, слово, язык, компания, вид, тип, жизнь, класс, область, данные, список, мир, основа, часть, право, приложение, безопасность, выполнение, исследование, действие, внимание, обеспечение, игра, зависимость, точка, разработка, имя, технология, форма, ресурс, обучение, доступ, место, отношение, параметр, путь, определение, требование, клиент, выбор, запрос, страна, сторона, инструмент, оценка, настройка, структура, материал, поле, связь, устройство, размер, обработка, группа} \\
\midrule
TR & the, and, bilgi, şekil, yer, zaman, durum, üzer, konu, sistem, önem, insan, değer, veri, hizmet, neden, alan, yol, yardımcı, eğitim, for, sahip, işlem, taraf, sonuç, özellik, uygulama, ifade, that, etki, hak, kontrol, ürün, örnek, yöntem, kaynak, gün, sağlık, servis, adım, hâl, dünya, dönem, öğrenci, kelime, tür, süreç, dikkat, sıra, kod, dil, yapı, yıl, dosya, müşteri, çocuk, süre, sayı, oyun, with, sorun, tarih, model, hareket, iletişim, teknoloji, kişi, analiz, kullanım, hedef, anlam, test, göz, program, soru, proje, içerik, ülke, hikaye, maliyet, yönetim, yaşam, nokta, cihaz, araç, bakım, bölge, ilişki, zeka, devam, açı, şey, hücre, web, kabul, faktör, typedef, çözüm, enerji, fonksiyon \\
\midrule
ZH &  \begin{CJK}{UTF8}{gbsn} 计算机, 稳定性, 重要性, 可靠性, 互联网, 优惠券, 机器人, 工程师, 影响力, 字符串, 一致性, 管理员, 注意力, 12月, 时间点, 参与者, 制造业, 可能性, 科学家, 年轻人, 实验室, 设计师, 吸引力, 房地产, 防火墙, 三角形, 多样性, 图书馆, 人民币, 专注于, 放大器, 建筑物, 收银台, 工作者, 研究者, 劳动者, 上下文, 显示器, 主持人, 积极性, 办公室, 障碍物, 艺术家, 俱乐部, 过滤器, 相关性, 爱好者, 糖尿病, 局限性, 真实性, 生产力, 抑郁症, 代表性, 自行车, 出版社, 半导体, 时间表, 攻击者, 分析师, 受害者, 所有人, 电动车, 阿拉伯语, 统计学, 蜂鸣器, 旅游业, 管理者, 使用者, 解码器, 独特性, 所有权, 百分比, 最小化, 研讨会, 相似性, 发动机, 学习者, 共和国, 委员会, 青少年, 无线电, 微生物, 原材料, 污染物, 排水沟, 连接器, 摄影师, 聚合物, 运动员, 农产品, 许可证, 格式化, 维生素, 11月, 所得税, 自信心, 流水线, 10月, 高血压, 吹风机 \end{CJK} \\
\midrule
DD & time, day, people, thing, lot, problem, way, year, thank, today, one, room, job, company, week, morning, work, kind, month, name, car, friend, food, night, idea, minute, price, hour, school, class, card, dollar, place, number, office, house, business, bus, course, afternoon, movie, tonight, city, country, bit, book, dinner, account, party, family, ticket, home, computer, phone, service, yuan, evening, girl, information, question, game, man, child, order, sale, life, seat, world, product, coffee, credit, moment, part, trip, help, restaurant, meeting, apartment, look, case, experience, plan, mind, doctor, guy, hand, form, parent, flight, water, student, news, table, size, position, boss, wife, music, line, weather \\
\midrule
Ubuntu & file, problem, thank, package, window, install, driver, command, thing, system, error, way, partition, server, time, card, user, drive, version, line, kernel, desktop, root, idea, boot, option, grub, network, one, issue, default, question, gnome, machine, screen, computer, program, terminal, stuff, bit, help, name, laptop, manager, disk, device, menu, script, folder, output, setting, source, video, link, hardware, lot, sound, type, module, work, message, bug, box, people, port, page, password, connection, software, upgrade, list, guy, man, mode, home, point, channel, key, case, permission, support, log, reason, flash, internet, tool, part, day, installation, client, monitor, process, access, application, image, wine, space, ram, interface, site \\
\bottomrule
\end{tabular}
\caption{Top-100 noun vocabularies per language and dataset, ranked by corpus frequency after WordNet synonym filtering.}
\label{tab:vocab}
\end{table*}

\end{document}